\documentclass[11pt]{article}
\usepackage{ijcnlp2011}
\usepackage{times}
\usepackage{url}
\usepackage{latexsym}

\usepackage{amsmath}
\usepackage{multirow}

\usepackage{float}
\usepackage{amsfonts}
\usepackage{amssymb}
\usepackage{graphicx}

\title{A Concise Query Language with Search and Transform Operations for Corpora with Multiple Levels of Annotation}

\author{Anil Kumar Singh\\
  School of Computer Engg.\\
  KIIT University, Bhubaneswar, India\\
  {\tt asinghfcs@kiit.ac.in}}

\date{}

\begin{document}
\maketitle
\begin{abstract}
  The usefulness of annotated corpora is greatly increased if there is an associated tool that can allow various kinds of operations to be performed in a simple way. Different kinds of annotation frameworks and many query languages for them have been proposed, including some to deal with multiple layers of annotation. We present here an easy to learn query language for a particular kind of annotation framework based on `threaded trees', which are somewhere between the complete order of a tree and the anarchy of a graph. Through `typed' threads, they can allow multiple levels of annotation in the same document. Our language has a simple, intuitive and concise syntax and high expressive power. It allows not only to search for complicated patterns with short queries but also allows data manipulation and specification of arbitrary return values. Many of the commonly used tasks that otherwise require writing programs, can be performed with one or more queries. We compare the language with some others and try to evaluate it.
\end{abstract}

\section{Introduction}
\label{intro}

Representation of annotated corpora and mechanisms to access and manipulate data have been a major area of research in Natural Language Processing (NLP) during the last many years. These are difficult problems (even when considering only text), primarily because linguistic annotation can be of various kinds and at multiple levels, e.g. morphological, Part Of Speech (POS) tagging, chunking, phrase structure, dependency relations, semantic relations, discourse and dialog information etc. Merging all such annotation~\cite{Suderman:06} for some corpus in one file per document is perhaps not possible or it may not be feasible. However, many of these annotation levels can indeed be merged in one file per document. One of the ways to do this is through a formalism based on threaded trees~\cite{JML:95,Ait-Mokhtar:02}, by having different kinds of threads for different kinds of annotation and putting constraints on the threads. The constraints ensure that the problem of `chasing pointers'~\cite{Bird:00a} does not become a serious problem.

One specific annotation scheme that uses threaded trees~\cite{Begum:08} encodes dependency trees and some other relations (such as co-reference) over chunked data by the use of constrained threads. It might be possible to translate many other annotation schemes into a threaded tree based scheme, but we leave that for future work, as here we will focus on the query language, not the annotation framework.

Though a lot of query languages have been proposed~\cite{Bird:00b,Lai:04}, we are not aware of any language for threaded tree based annotated data. Languages for linguistic trees have, however, been well studied and we will try to relate our language with a few of those.

We will first present a short review of the related work (Section-\ref{related-work}). Then we briefly discuss how threaded trees can be used for encoding multiple layers of annotation (Section-\ref{threaded-trees}). After discussing the requirements of a query language for searching and transforming annotated data in Section-\ref{requirements}, we present a brief overview of the syntax of the language in Section-\ref{syntax}, followed by a description of the syntactic elements in Section-\ref{syntactic-elements}. We then suggest some applications of the language (Section-\ref{apps}), before presenting a comparative assessment of the language and its limitations (Section-\ref{comparative-assessment}). We also derive some directions for future work based on this.


\section{Related Work}
\label{related-work}

In this section we present a short review of related work reported in the literature under two headings: 1) annotation frameworks, and 2) query languages.

\subsection{Annotation Frameworks}
\label{annotation-frameworks}

Annotation frameworks can be divided into two broad categories: graph based and tree based. A comparative study of many annotation frameworks was presented by Bird and Liberman~\shortcite{Bird:98,Bird:00a}. Since their work was more focused on speech data, many of the frameworks considered were meant for such data, e.g. TIMIT, CHILDES and MATE. But they also considered some frameworks which are used more for text based data, such as the Penn Treebank corpus. One major framework that was not included in their study was the GATE annotation framework~\cite{Cun02b}, which uses standoff format (a common way to allow multiple layers of annotation). After considering each of them, they proposed a formal framework for linguistic data that could be used for all those purposes for which these frameworks are used. They called the proposed framework an `annotation graph'. The nodes in this graph were temporal points, while the edges represented the linguistic objects. They showed how even multiple layers of annotation could be represented by different tiers of the annotation graph. Maeda et al.~\shortcite{Maeda:02} described how the Annotation Graph Toolkit could be used for creating tools for this framework.

As compared to graph based annotation frameworks, tree based frameworks are (for obvious reasons) much simpler. However, adding extra layers of annotation to tree based data is a non-trivial problem. One way is to store such extra information in a separate file and use node identifiers to link it to the data in the tree. Another way is to store the data in multiple trees stored in different files, which are somehow linked together. Yet another way is to use threaded trees, which is the one we will be assuming for our query language.

In one of the major works, Cotton and Bird~\shortcite{Cotton:02} had proposed an integrated framework for treebanks and multilayer annotations. This work focused more on tree based data, but it also suggested annotation graphs as the solution.

\subsection{Query Languages}
\label{corpus-query-languages}

Bird et al.~\shortcite{Bird:00b} had compared some of the query languages available (at that time) for graph based annotation frameworks. These included Emu and the MATE query language. They then proposed their own query language for annotation graphs. This language used path patterns and abbreviatory devices to provide a convenient way to express a wide range of queries. This language also exploited the quasi-linearity of annotation graphs by partitioning the precedence relation to allow efficient temporal indexing of the graphs. Another such survey was by Lai and Bird~\shortcite{Lai:04}, where the authors considered TigerSearch, CorpusSearch, NiteQL, Tgrep2, Emu and LPath~\cite{Bird:05,Bird:06b}. From this study, the authors tried to derive the requirements that a good tree query language should satisfy.

\begin{figure*}[tb]
\label{fig:tangled-tree-rep}
\centering
\includegraphics[scale=0.65]{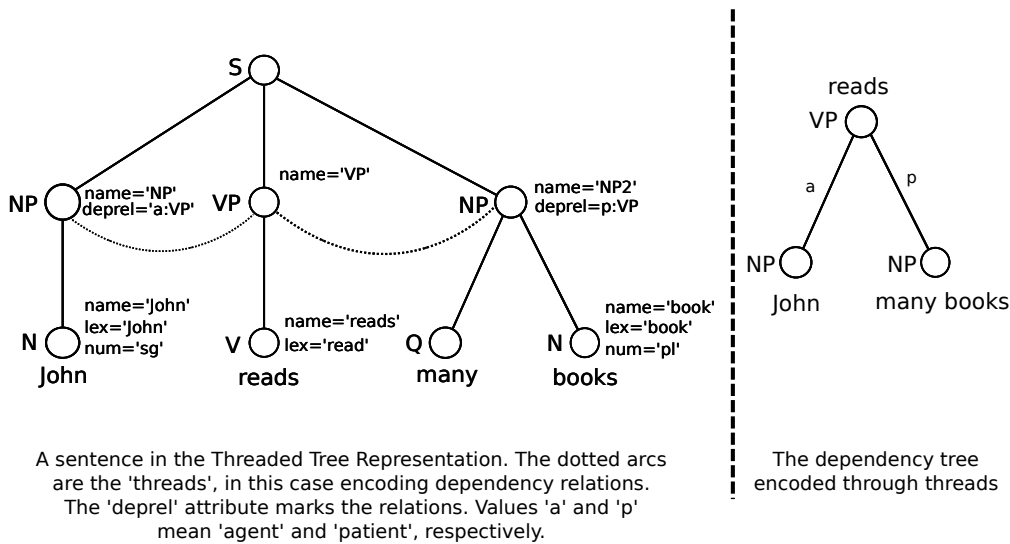}
\caption{Threaded Tree Representation}
\end{figure*}

Resnik and Elkiss~\shortcite{Resnik:05} had reported a search engine for linguists that was meant to be easy to use for linguists who were not versed in the use of computers. This tool allowed linguists to draw patterns in the form of sub-trees, which were then converted into queries and searched. Like almost all such languages, it did not allow manipulation of data and it worked only for certain levels of annotation. It was mainly aimed at searching  phrase structure patterns and morphological information.

One of the well known query languages for annotated corpora used for linguistic studies and for NLP is the Corpus Query Language\footnote{\url{http://www.fi.muni.cz/~thomas/corpora/CQL/}} (CQL). It is used in a popular tool called Sketch Engine\footnote{\url{ http://www.sketchengine.co.uk/}}~\cite{Kilgarriff:04}. It provides a wide variety of functionalities to access corpora, such as searching words, lemmas, roots, POS tags of a word, getting the left and right contexts upto a window size of 15.


Another usual practice is to have a query tool for syntactically annotated corpora such that the data is converted internally to relational database and the query is written using SQL~\cite{Kallmeyer:00}. A much earlier work was titled `A modular and flexible architecture for an integrated corpus query system'~\cite{Christ:94}, which is used by the IMS Corpus Workbench\footnote{\url{http://www.ims.uni-stuttgart.de/projekte/CorpusWorkbench/}}. Another query language called MQL is used in the Emdros database engine for analyzed or annotated text\footnote{\url{http://emdros.org/mql.html}}. MQL is a descendant of QL~\cite{Doedens:94}.

The language that we describe here is similar in some aspects to many of these languages, but different in others. The most important differences are the support for threaded trees, its very concise syntax, query-and-action mechanism (data manipulation), arbitrary return values, support for custom commands and the possibility for pipelining results through the source and destination operators. It also has high expressive power generally. Moreover, it can be used for purposes other than NLP because the data that it operates on is similar to the general XML representation and the language has {\bf some} of the power of both XPath\footnote{\url{http://www.w3.org/TR/xpath}} based querying and XSLT\footnote{\url{http://www.w3.org/TR/xslt}} based transformation.

\section{Threaded Trees and Multiple Layers of Annotation}
\label{threaded-trees}

There can be a different kind of `annotation graph' where nodes are the linguistic objects and edges are relations (which could include hierarchical relations such parent-child or dominated-by). Threaded trees~\cite{JML:95,Ait-Mokhtar:02} are a subset of this kind of annotation graphs. The base structure is a tree and threads are the crossing edges. If threads are typed and labelled, then each type can be used to represent one layer of annotation. The labels can be used to indicate annotated relations of that type. Depending on the requirements of the annotation framework, strong constraints can be applied to the threads. For example, we could have a constraint that says that the threads of a particular type are required to form a tree such that it has all the leaf nodes that the base tree has. Such threads can then be used to represent, say, dependency relations, assuming that the base tree represents the phrase structure (or just a POS tagged and chunked sentence). Other threads may have different kind of constraints and they can be used to encode, say, the argument structures or semantic relations.

By allowing the nodes to have feature structures associated with them, we can have deeper annotation of the properties of nodes as well as relations. In fact, threads themselves may be marked via the feature structures. We will assume here that these feature structures are sets of attribute-value pairs, but in an extended version, the values could be feature structures, thus allowing nested feature structures.

The Corpus Query Language that we present here assumes that the data is in the above representation (Figure 1). The language allows the data to be searched, manipulated and extracted, irrespective of how many levels of annotations are stored in the threaded tree structure.

Threaded trees can be easily implemented using XML as the data format.

\section{Requirements of a Search and Transform Language}
\label{requirements}

Lai and Bird~\shortcite{Lai:04} had identified some requirements that a good tree query language should satisfy (in terms of expressive power). These can be summarized under four headings:

\begin{itemize}
\item {\bf Tree Navigation}: This will allow operations for subtree matching, returning subtrees, reverse navigation (e.g. `follows') and non-tree navigation (e.g. over terminals).
\item {\bf Closures}: Of basic relations such as dominance, precedence and sibling precedence, as well as of more complex relations such as self-recursive rules using a closure or closures involving more than one step, i.e., things which are expressible using XPath.
\item {\bf Beyond-Tree Navigation}: For searching beyond sentence boundaries or for searching over forests. Also, for querying non-tree structures (the threads in our case might form a non-tree structure).
\item {\bf Update}: Operations like insertion, deletion, moving and labelling of nodes, subject to the constraint that the base text is preserved.
\end{itemize}

Apart from these specific requirements, there are some general requirements that any programming or scripting language should satisfy. These include conciseness of the syntax, short learning curve, scope for efficient implementation and possibility of translation to other popular languages. In the following sections and especially in Section-\ref{comparative-assessment}, we will discuss the degree to which our language meets these requirements.

\begin{table*}[!ht]
\label{table:syntacic-elements}
\small
\centering
\begin{tabular}{|l|l|}
\hline
\multicolumn{2}{|c|}{{\bf Objects}}\\
\hline
\hline
{\bf Object}	& {\bf Remarks}\\
\hline
{\bf F} and {\bf S}	& The File (document) and the Sentence\\
{\bf C}	& The Current node (the centre of the node-by-node query processing)\\
{\bf P} and {\bf N}	& The Previous and the Next nodes\\
{\bf Pr} and {\bf Nx}	& The Previous sibling and the Next sibling nodes\\
{\bf A} and {\bf D}	& The Ancestor (or parent) and the Descendant (or child) nodes\\
{\bf R} and {\bf T}	& The Referred and the Referring nodes (thread navigation)\\
{\bf M}	& The node(s) that matched one of the conditions, e.g. M[p], p being the condition alias\\
\hline
\hline
\multicolumn{2}{|c|}{{\bf Members}}\\
\hline
\hline
{\bf Member}	& {\bf Remarks}\\
\hline
{\bf l}	& The lexical data for the node\\
{\bf t}	& The tag (e.g. POS tag) of the node\\
{\bf a}	& The attribute, with the index specified within square brackets, e.g. a[`lex']\\
{\bf v}	& The level (distance from the root) of a tree node (0 being the root)\\
{\bf f}	& Boolean value to check if a node is a leaf node\\
\hline
\hline
\multicolumn{2}{|c|}{{\bf Operators and Values}}\\
\hline
\hline
{\bf Operator/Value}	& {\bf Remarks}\\
\hline
{\bf AND}	& Conjunction of two or more search conditions\\
{\bf OR}	& Disjunction of two or more search conditions\\
{\bf ()}	& Parenthesis: Grouping of search conditions for evaluation or nesting\\
{\bf []}	& Index: Integer (position) or string (name or alias), e.g. D[2], a[`deprel'] etc.\\
{\bf :}	& Index qualifier, e.g. D[2:3] (grandchild's third child)\\
{\bf .}	& The dot operator to access the members of an object\\
	& and to form node addresses\\
{\bf `'}	& The literal value specified within single quotes (e.g. `agent'),\\
	& usually of members\\
{\bf +}	& Concatenation: To join together two or more literal values or variables\\
{\bf = and !=}	& Equal and Not Equal (LHS), based on exact equality of values\\
{\bf $\sim$ and !$\sim$}	& Similar and Not Similar (LHS), based on similarity, e.g. using regex\\
{\bf =}	& Value assignment operator (RHS)\\
{\bf --$>$}	& Action to be performed on the nodes that matched the conditions\\
{\bf =:}	& The sources of the data, e.g. the corpus files\\
{\bf :=}	& The destinations, e.g. the files where the results have to be stored\\
{\bf /}	& Alias assignment for conditions, return values and sources/destinations\\
\hline
\hline
\multicolumn{2}{|c|}{{\bf Wildcards and Ranges}}\\
\hline
\hline
{\bf Wildcard/Range}	& {\bf Remarks}\\
\hline
{\bf ?}	& The first node to match\\
{\bf .}	& The last node to match\\
{\bf *}	& Any nodes to match (disjunction)\\
{\bf @} & All node(s) that match(es), e.g. N[@], M[@] (conjunction)\\
{\bf 0} & None (normal indices start from 1)\\
{\bf --}	& The range of nodes, e.g. N[2-4], P[3-], D[-2] etc. and z is the last node.\\
\hline
\end{tabular}\\[2pt]
\caption[A1]{A Summary of the Query Language}
\end{table*}

\section{An Overview of the Syntax}
\label{syntax}

One way to express the motivation for developing this is as follows. Suppose there was an exhaustive API (Application Programming Interface) to process, search and manipulate the data in the Threaded Tree Representation. This API allows all kinds of possible operations on the data. Then the proposed query language should be able to allow the same range of operations on this kind of data, just by providing concise queries. What we present in this document is the first draft of such a language, so it does not cover this whole range of operations, but it does cover a fairly large part which would be the most useful for developers as well as other users.


A query in this language can consist of four parts, out of which only the second part (conditions) is mandatory:

\begin{enumerate}
 \item {\bf Sources (src)}: The documents or data streams on which the query has to be executed. (Optional)
 \item {\bf Conditions (cnd)}: The search conditions based on which the values will be returned and the actions (if any) will be taken
 \item {\bf Actions (act)}: The data manipulation operations which have to be performed on the nodes which matched the search conditions. The operations could (optionally) simply specify return values. (Optional)
 \item {\bf Destinations (dst)}: The documents or data streams to which the results have to be stored or transferred. (Optional)
\end{enumerate}

The the top level description of the syntax of a query would be:

\begin{equation}
 [src~=:]~cnd~[->~act]~[:=~dst]
\end{equation}

Table 1 presents a summary of the syntactic elements of the language (Section-\ref{syntactic-elements}). An simple example of a query is:

\small
\begin{verbatim}
C.t='NN' -> C.t='Noun' and A
\end{verbatim}
\normalsize

The above query has only two parts: the conditions (provided on the Left Hand Side or LHS) and the actions (provided on the Right Hand Side or RHS). This query will replace all the `NN' tags of the tree nodes in the document with the `Noun' tag and return the parent node (the first ancestor, A[1] or just A). {\bf C} here represents the current node, {\bf t} represents the tag, {\bf .} is the dot operator, {\bf =} is the value assignment operator and {\bf --$>$} is the action operator. If we want the action to be applied only on the leaf nodes (which will usually hold the actual tokens with some lexical data), we can write:

\small
\begin{verbatim}
C.t='NN' and C.f='t' -> C.t='Noun'
\end{verbatim}
\normalsize

Note that we are not following any specific linguistic formalism here and the examples are only to demonstrate the syntax of the language. It is up to the users of the language to write linguistically significant queries for their own specific purposes.

The previous two queries use one of the logical operators (and/or) and the literal value 't', which means 'true'. If we add the source and destination to the query (leaving out the terminal node condition to save space), it will be:

\small
\begin{verbatim}
xml:src.txt:UTF-8 =: C.t='NN' \
  -> C.t='Noun' := xml:tgt.txt:UTF-8
\end{verbatim}
\normalsize

Two more operators are introduced in the above query: the source ({\bf =:}) and the destination ({\bf :=} operators). If, instead of an action, we want the query to return the current, the previous and the next node, then the query would be:

\small
\begin{verbatim}
xml:src.txt:UTF-8 =: C.t='NN' \
  -> C and P and N := raw:tgt.txt:UTF-8
\end{verbatim}
\normalsize

How the result is displayed or stored will depend on the format specified (raw, i.e., simple text, in the above query and xml in the preceding query) as well as on the implementation, e.g. how exactly the multiple values are added to the destination. The current version of the language does provide some control for this through the process of concatenation:

\small
\begin{verbatim}
C.t='NN'-> C.l+'-'+C.t+'; \
  '+P.l+'-'+P.t';'+N.l+'-'+N.t';'/r
\end{verbatim}
\normalsize

The alias ({\bf /}) operator here assigns an alias (a name or a key), viz. r, to the return value. Another new operator above is the concatenation ({\bf +}) operator.

The present implementation will put the concatenation of values on the RHS of the above query on one line (assuming simple text output), preceded by the alias for the return value (if given) or the query term representing the return value followed by `: '. Alternatively, we could write:

\small
\begin{verbatim}
C.t='NN'-> C.l+'-'+C.t/c \
  and P.l+'-'+P.t/p and N.l+'-'+N.t/n
\end{verbatim}
\normalsize

For the above query, the current implementation would put the three return values on three separate lines, each preceded by the respective alias followed by `: '.

One very important feature is that the LHS can be prefixed by a directive such as ``TT['deprel']:'' that will `extract' the tree encoded by the thread of type `deprel' and then all the usual tree operations that can be performed on the base tree can also be performed on this tree (in this case a dependency tree):

\small
\begin{verbatim}
TT['deprel']: C.t='NP' AND A.t='VP'
\end{verbatim}
\normalsize


\section{Syntactic Elements}
\label{syntactic-elements}

The syntactic elements of the language can be divided into the following categories: objects and members, operators and values, wildcards and ranges, source and destinations, actions and return values, and custom commands. Together, they provide the expressive power that allows the language to fulfil most of the requirements mentioned in Section-\ref{requirements}.

\subsection{Objects and Members}

Objects are the tree nodes or larger (e.g. S or sentence and F or file/document) units of linguistic data on which the queries operate. Query processing (in most cases) happens node-by-node and the centre of this processing is the C or the Current node. For navigation, all the node addresses are based on the current node. Thus, we can have a previous node (P), a next node (N), an ancestor node (A), a descendant node (D), a referred node (R) and a referring node (T), the last two meant for threads. While P is for precedence, there is corresponding node Pr for sibling precedence. Similarly, there is a corresponding node Nx for N. There is another special node (M) to represent the nodes that matched a query condition. Most of the node types (except C) can have multiple candidates for matching, which are specified through integer indices (for P, N, Pr, Nx and D, e.g. N[2] for the next to next node), through string keys (for R, e.g. R['deprel']) or through a string key and an integer key (for T, e.g. T['deprel':2]). The indices are enclosed inside square brackets. The string keys are enclosed in single quotes, except when the key is an alias because aliases are not evaluated whereas other keys can be evaluated as they can consist of variables. Any value (including node indices) is treated as a variable and is evaluated if it is not enclosed within single quotes.

Compound node addresses can be created using the dot operator, e.g. N.A[2].P[3], which will mean the preceding node at a distance of 3 (P[3]) of the grandparent node (A[2]) of the next node (N). Variables can be either node addresses, or member values (e.g. N.t) or a combination of variables and literals formed by using the concatenation operator (e.g. N.t+'-'+N.l).

The four members currently implemented are: {\bf l} or the lexical data, {\bf t} or the tag, {\bf v} or the level in the tree and the boolean member {\bf f} to check if a node is a leaf node. While the object symbols are written in capital letters, the members symbols are written in small letters.

\subsection{Operators and Values}

Some of these have already been introduced in the preceding section. Here we will add some more information about them. The dot operator ({\bf .}) allows us to form node addresses or to access the members of the nodes. The comparison operators (used on the LHS) currently provided are: equality ({\bf =}), inequality ({\bf !=}), similarity ({\bf $\sim$}) and not-similarity ({\bf !$\sim$}). The value assignment operator reuses the symbol  ({\bf =}) on the RHS. The LHS and the RHS are separated by the action operator ({\bf --$>$}). The alias assignment operator ({\bf /}) provides an easy way to write concise queries because we can give single character aliases, apart from allowing (more readable) access to objects previously mentioned in the query. The concatenation operator ({\bf +}) has already been explained. Parentheses ({\bf (\dots)}) are used to group together query conditions for prioritized evaluation and to form nested queries. The logical operators AND, OR and NOT (written as !(\ldots)) operators are also supported to form complex queries.

\subsection{Wildcards and Ranges}

In many cases, it can be very useful to be able to specify wildcards when there can be multiple candidates (the closure requirement of Section-\ref{requirements}). The language provides three wildcards and also ranges in terms of integer indices:

\begin{itemize}
 \item {\bf ?}: The first one to match, e.g. N[?]
 \item {\bf .}: The last one to match, e.g. N[.]
 \item {\bf *}: Any node(s) that match(es), e.g. N[*], M[*] (disjunction)
 \item {\bf @}: All node(s) that match(es), e.g. N[@], M[@] (conjunction)
 \item {\bf 0}: None (normal indices start from 1)
 \item {\bf -}: From the first index to the second index, e.g. P[2-4], P[2-], P[-2]. The last node is specified by the special index z.
\end{itemize}

Here is one example of using wildcards, aliases and concatenation:

\small
\begin{verbatim}
P[*].t/p='XC' and C.t!='XC' \
 -> M[p:*].t=C.t+'C'
\end{verbatim}
\normalsize

Suppose there are tag sequences of the form XC XC NN, XC XC JJ etc. and they have to be converted to NNC NNC NN and JJC JJC JJ, respectively, then the above query will do that.

\subsection{Sources and Destinations}

In the current implementation, the sources and destinations can only be files, but they could, in principle, be streams too, e.g. for building pipelines of queries. In the case of files, a source or a destination can be specified in terms of four parts:

\begin{itemize}
 \item {\bf Format}: Could be simple text or XML or something else, depending on the implementation
 \item {\bf Location}: The URL or the URI or the path of the document or file
 \item {\bf Charset}: The charset or encoding of the document (the default is UTF-8)
 \item {\bf Name}: The object alias
\end{itemize}

An example of a source specification is:

\begin{verbatim}
xml:src.txt:UTF-8/s
\end{verbatim}

In the above query, xml is the format, src.txt is the location, UTF-8 is the charset and $s$ is the name or the alias. Aliases allow multiple sources and destinations to be specified and also accessed from other parts of the query. For example the following query uses two sources:

\small
\begin{verbatim}
xml:src1.txt:UTF-8/s1 \
  and xml:src2.txt:UTF-8/s2 \
  =: F[s1].C.t='NN' and F[s2].C.t='Noun'
\end{verbatim}
\normalsize


\subsection{Actions and Return Values}

Multiple transformations can be performed by using the AND operator on the RHS (Right Hand Side), including on nodes other than the current node by using the same notation as for the LHS.

On the RHS, we can also specify return values by using the node symbols and the dot notation (e.g., $C$, $N.A$). For this purpose, another symbol S can be used to return sentences for the nodes which match. The syntax is intuitive and easy to remember. If we don't provide an assignment value, then the node address, variable or the concatenated value is treated as the return value (e.g. $N.l$ will return the next node's lexical data, whereas $N$ will return the next node).

If an assignment expression has nodes on both sides, it will be interpreted as a node insertion, deletion or move operation. For example, $Nx=A.N.D$ will take the next sibling node and move it so that it is dominated by the node which is next to the parent of the current node. At present there is no way to ensure that base text is preserved (the user is expected to ensure that), but we will introduce a mechanism for this in a future version.

\subsection{Creating and Navigating Threads}

Threads (which represent multiple layers of annotation) can be created by providing a query like the following:

\small
\begin{verbatim}
C.l='reads' AND C.f='t' \
   AND A.N[?].t/q='NP' \
  -> M[q].a['deprel']=''a':A.a['name']'
\end{verbatim}
\normalsize

This query will look for a leaf node with the lexical data 'reads' such that its parent node is followed by an NP (the first one encountered). Then, according to the action specified on the RHS, it will create a thread from this NP to the parent node (VP) of 'reads' by adding a value for the $deprel$ attribute as a concatenation of the relation label ('a' or agent) and the unique name of the VP, separated by a colon.

To navigate threads, the node symbols R (referred node) and T (referring node) can be used. For the dependency tree in the example given in Figure 1, VP is the referred node and the two NPs are the referring nodes. Thus the following query:

\small
\begin{verbatim}
C.t='NP' AND R['deprel'].t='VP'
\end{verbatim}
\normalsize

will find all the NPs for which the referred node is a VP. Note that we may have to specify the attribute used for the kind of thread we are searching. The most commonly used of these attributes could be used as a default, so that there is no need to specify it. Since there can be more than one referring nodes, we need to specify the index as well in the case of such nodes:

\small
\begin{verbatim}
C.t='VP' AND T['deprel':2].t='NP'
\end{verbatim}
\normalsize

The above query will search for all the VPs whose second referring node (e.g. the second argument) is an NP.

\subsection{Commands}

The language also allows us to specify commands to be executed on the data. For example, if we want to ensure that all the nodes have unique names before we start providing transformation actions, we can give a command like:

\small
\begin{verbatim}
reallocateNames
\end{verbatim}
\normalsize

where `names' are the unique node identifiers which are used for marking and navigating threads.

Note that the language supports custom commands, so there is no exhaustive list of commands. The current implementation has some commands that have been found useful so far, but more can be easily added. Commands can also be executed subject to some conditions, if we write a query with LHS giving the conditions and the RHS giving the command:

\small
\begin{verbatim}
C.a['name']='' -> reallocateNames
\end{verbatim}
\normalsize

This query will ensure that the nodes have unique names when we start marking threads for any kind of extra annotation layer.



\section{Applications}
\label{apps}

Only a few examples have been given in the preceding sections. The query language allows many other kinds of operations using the constructs that have been mentioned. These operations can be used for many purposes, apart from searching. Queries can be written to perform sanity checks on the annotated data. They can be used to automatically mark information that is very predictable (to reduce manual work). They can be used to bring the old annotated data in tune with the new annotation specifications without having to write programs for that purpose, even if the changes require more complex operations than simple global replacement, as the XC XC NN example given above shows. Queries can also be written to easily extract complex features for, say, machine learning algorithms. They can be used to make the task of an annotation adjudicator easier.


\section{Comparative Assessment}
\label{comparative-assessment}

Bird and Lai~\shortcite{Lai:04} had used seven syntactic queries to compare various tree query languages. These are given in Table 2. The evaluation criterion was at least two fold. First, how many of these queries are expressible in a language. Second, how concise are those queries.

\begin{table}[tb]
\label{table:eval-queries}
\small
\begin{tabular}{ll}
Q1. & Find sentences that include the word `saw'.\\
Q2. & Find sentences that do not include the word `saw'.\\
Q3. & Find noun phrases whose rightmost child is a noun.\\
Q4. & Find verb phrases that contain a verb immediately\\
    & followed by a noun phrase that is immediately\\
    & followed by a prepositional phrase.\\
Q5. & Find the first common ancestor of sequences\\
    & of a noun phrase followed by a verb phrase.\\
Q6. & Find a noun phrase which dominates a word `dark'\\
    & that is dominated by an intermediate phrase\\
    & that bears an L-tone.\\
Q7. & Find an noun phrase dominated by a verb phrase.\\
    & Return the subtree dominated by that noun\\
    & phrase only.\\
\end{tabular}
\caption{Syntactic Queries for Comparing Tree Query Languages (Bird and Lai, 2004)}
\end{table}

All seven of these queries can be expressed in our language, but a feature required for the fifth query (`@' index for all nodes to match: conjunction, rather than disjunction, as expressed by the `*' index) has not yet been fully implemented. These queries are given in Table 3.

\begin{table}[tb]
\label{table:eval-queries-scql}
\small
\begin{tabular}{ll}
Q1. & C.l='saw' $->$ S\\
Q2. & C.l='0' AND C.D[*:0].l/p='saw' $->$ S\\
Q3. & C.t='NP' AND C.D[z].t='NN'\\
Q4. & C.t='VP' AND C.D[*].t~'V*'/p\\
    & AND M[p].N.t='NP' AND M[p].N[2].t='NP'\\
Q5. & P[*].t/p='NP' and C.t='VP' AND\\
    & M[p:@].A[*]=C.A[*]/q $->$ M[q:1]\\
Q6. & C.t='NP' AND D[*].l='dark'/p\\
    & AND M[p].A[*].a['tone']='LTone'/q\\
    & AND C.l$>$M[q].l\\
Q7. & C.t='NP' AND C.A[*].t='VP'\\
\end{tabular}
\caption{Comparison Queries in Our Language}
\end{table}

Only NiteQL could express all these queries, others (TigerSearch, Emu, CorpusSearch, Tgrep2 and LPath) could not express at least one of these seven queries. However, NiteQL does not have a concise syntax. Thus, in terms of expressive power, our language compares favourably with these languages. It also has comparable conciseness of syntax. Moreover, there are a range of queries that cannot be expressed in other languages because they are not meant for threaded trees and have no equivalents to the R and T nodes that we have.

But a look at the queries in Table 3 also shows some scope for improvement. For example, there is need for `dominates' and `is-dominated-by' operators which will make these queries even more concise. As our language was developed initially for annotators working on data that was not huge in quantity, there seem to be some problems from the efficiency point of view too.

Other directions for future work include a study of the formal properties of the language and a more rigorous evaluation based on various criteria. We are also developing a graphical query designer for this language.

Even in the present condition, the language is being used by many annotators and annotation adjudicators to make their work easier, as well as by a few developers to build substitutes for previously used programs for tasks like sanity checks, validation etc. Users with non-computational background have found it easy to learn at least the basics of it, though we have not yet performed a proper evaluation of the ease of learning. Our non-empirical experience is that power users can learn it within a few hours, or even less in some cases.

\section{Conclusion}

We presented a concise yet expressive query language for data that is in a tree-like format such that one node can have links (`threads') to any other node in the tree, allowing for additional trees or graphs to be encoded in the core tree. The language uses simple elements and constructs like objects, members, operators, variables, values, nesting, aliases, wildcards etc. to allow writing queries that can perform fairly complex operations without the need to write programs for this purpose. Multiple conditions can be given, multiple transformation actions can be specified, multiple return values can be specified and so can be multiple sources and destination. The language can, in fact, be used as a scripting language for annotated data with multiple levels of annotation, where multiple levels are encoded through `typed' threads. We compared the language to some other query languages and suggested some directions for future work.


\bibliographystyle{acl}
\bibliography{biblio}  

\begin{thebibliography}{}

\bibitem[\protect\citename{Ait-Mokhtar \bgroup et al.\egroup
  }2002]{Ait-Mokhtar:02}
S.~Ait-Mokhtar, J.P. Chanod, and C.~Roux.
\newblock 2002.
\newblock Robustness beyond shallowness: incremental deep parsing.
\newblock {\em Natural Language Engineering}, 8(2-3):121--144, January.

\bibitem[\protect\citename{Begum \bgroup et al.\egroup }2008]{Begum:08}
R.~Begum, S.~Husain, A.~Dhwaj, D.M. Sharma, L.~Bai, and R.~Sangal.
\newblock 2008.
\newblock {Dependency annotation scheme for Indian languages}.
\newblock {\em Proceedings of IJCNLP-2008}.

\bibitem[\protect\citename{Bird and Liberman}1998]{Bird:98}
Steven Bird and Mark Liberman.
\newblock 1998.
\newblock Towards a formal framework for linguistic annotations.
\newblock In {\em Proceedings of the International Conference on Spoken
  Language Processing}, Sydney.

\bibitem[\protect\citename{Bird and Liberman}2000]{Bird:00a}
Steven Bird and Mark Liberman.
\newblock 2000.
\newblock A formal framework for linguistic annotation.
\newblock {\em Speech Communication}, 33(1,2):23--60.

\bibitem[\protect\citename{Bird \bgroup et al.\egroup }2000]{Bird:00b}
Steven Bird, Peter Buneman, and Wang-Chiew Tan.
\newblock 2000.
\newblock Towards a query language for annotation graphs.
\newblock In {\em Proceedings of the Second International Conference on
  Language Resources and Evaluation}, pages 807--814, Athens, Greece.

\bibitem[\protect\citename{Bird \bgroup et al.\egroup }2005]{Bird:05}
Steven Bird, Yi~Chen, Susan Davidson, Haejoong Lee, and Yifeng Zheng.
\newblock 2005.
\newblock Extending xpath to support linguistic queries.
\newblock In {\em Proceedings of Programming Language Technologies for XML},
  pages 35--46, Long Beach, California. Association for Computing Machinery.

\bibitem[\protect\citename{Bird \bgroup et al.\egroup }2006]{Bird:06b}
Steven Bird, Yi~Chen, Susan Davidson, Haejoong Lee, and Yifeng Zheng.
\newblock 2006.
\newblock Designing and evaluating an xpath dialect for linguistic queries.
\newblock In {\em Proceedings of the 22nd International Conference on Data
  Engineering}, pages 52--61, Atlanta, USA.

\bibitem[\protect\citename{Christ}1994]{Christ:94}
Oli Christ.
\newblock 1994.
\newblock A modular and flexible architecture for an integrated corpus query
  system.
\newblock In {\em Proceedings of COMPLEX'94}, Budapest.

\bibitem[\protect\citename{Cotton and Bird}2002]{Cotton:02}
Scott Cotton and Steven Bird.
\newblock 2002.
\newblock An integrated framework for treebanks and multilayer annotations.
\newblock In {\em Proceedings of the Third International Conference on Language
  Resources and Evaluation}, pages 1670--1677, Las Palmas, Spain.

\bibitem[\protect\citename{Cunningham \bgroup et al.\egroup }2002]{Cun02b}
H.~Cunningham, D.~Maynard, K.~Bontcheva, and V.~Tablan.
\newblock 2002.
\newblock {GATE: A framework and graphical development environment for robust
  NLP tools and applications}.
\newblock In {\em Proceedings of the 40th Anniversary Meeting of the ACL}.

\bibitem[\protect\citename{Doedens}1994]{Doedens:94}
Crist-Jan Doedens.
\newblock 1994.
\newblock {\em Text Databases. One Database Model and Several Retrieval
  Languages (Language and Computers)}.
\newblock Editions Rodopi Amsterdam, Amsterdam and Atlanta.

\bibitem[\protect\citename{Kallmeyer}2000]{Kallmeyer:00}
Laura Kallmeyer.
\newblock 2000.
\newblock A query tool for syntactically annotated corpora.
\newblock In {\em Proceedings of the 2000 Joint SIGDAT conference on Empirical
  methods in natural language processing and very large corpora}, pages
  190--198.

\bibitem[\protect\citename{Kilgarriff \bgroup et al.\egroup
  }2004]{Kilgarriff:04}
Adam Kilgarriff, Pavel Rychly, Pavel Smrz, and David Tugwell.
\newblock 2004.
\newblock A query tool for syntactically annotated corpora.
\newblock In {\em Proceedings of Euralex}, pages 105--116.

\bibitem[\protect\citename{Lai and Bird}2004]{Lai:04}
Catherine Lai and Steven Bird.
\newblock 2004.
\newblock Querying and updating treebanks: A critical survey and requirements
  analysis.
\newblock In {\em Proceedings of the Australasian Language Technology
  Workshop}, pages 139--146, Sydney, Australia.

\bibitem[\protect\citename{Larchevêque}1995]{JML:95}
J.M. Larchevêque.
\newblock 1995.
\newblock Optimal incremental parsing.
\newblock {\em ACM Transactions on Programing Languages and Systems},
  17(1):1--15, January.

\bibitem[\protect\citename{Maeda \bgroup et al.\egroup }2002]{Maeda:02}
Kazuaki Maeda, Steven Bird, Xiaoyi Ma, and Haejoong Lee.
\newblock 2002.
\newblock Creating annotation tools with the annotation graph toolkit.
\newblock In {\em Proceedings of the Third International Conference on Language
  Resources and Evaluation}, pages 1914--1921, Las Palmas, Spain.

\bibitem[\protect\citename{Resnik and Elkiss}2005]{Resnik:05}
Philip Resnik and Aaron Elkiss.
\newblock 2005.
\newblock The linguist's search engine: an overview.
\newblock In {\em ACL '05: Proceedings of the ACL 2005 on Interactive poster
  and demonstration sessions}, pages 33--36, Morristown, NJ, USA. Association
  for Computational Linguistics.

\bibitem[\protect\citename{Suderman and Ide}2006]{Suderman:06}
Keith Suderman and Nancy Ide.
\newblock 2006.
\newblock Layering and merging linguistic annotations.
\newblock In {\em Proceedings of the 5th Workshop on NLP and XML:
  Multi-Dimensional Markup in Natural Language Processing}, pages 89--92,
  Trento, Italy.

\end{thebibliography}


\end{document}